\documentclass[10pt, a4paper]{article}

\usepackage[final]{lrec2026}
\usepackage{times}
\usepackage{latexsym}
\usepackage[T1]{fontenc}

\usepackage[utf8]{inputenc}

\usepackage{microtype}

\usepackage{inconsolata}

\usepackage{graphicx}
\usepackage[table]{xcolor}
\usepackage{mathptmx}
\usepackage{booktabs}
\usepackage{multirow}
\usepackage{array}
\usepackage{float}
\usepackage{siunitx}
\usepackage{threeparttable}
\usepackage{tabularx}
\title{The Riddle of Reflection: Evaluating Reasoning and Self-Awareness in Multilingual LLMs using Indian Riddles}

\name{Abhinav P M\textsuperscript{\normalfont\sffamily 1}, 
       Ojasva Saxena\textsuperscript{\normalfont\sffamily 2}, 
       Oswald C\textsuperscript{\normalfont\sffamily 3}, 
       Parameswari Krishnamurthy\textsuperscript{\normalfont\sffamily 1}}

\address{\textsuperscript{1}International Institute of Information Technology, Hyderabad \\
         \textsuperscript{2}Independent Researcher \\
         \textsuperscript{3}National Institute of Technology, Tiruchirappalli \\
         {abhinav.pm@research.iiit.ac.in, ojassaxena03@gmail.com,} \\
         {oswald@nitt.edu, param.krishna@iiit.ac.in}}

\abstract{
The extent to which large language models (LLMs) can perform culturally grounded reasoning across non-English languages remains underexplored. This paper examines the reasoning and self-assessment abilities of LLMs across seven major Indian languages-Bengali, Gujarati, Hindi, Kannada, Malayalam, Tamil, and Telugu. We introduce a multilingual riddle dataset combining traditional riddles with context-reconstructed variants and evaluate five LLMs-Gemini 2.5 Pro, Gemini 2.5 Flash, Mistral-Saba, LLaMA 4 Scout, and LLaMA 4 Maverick-under seven prompting strategies. In the first stage, we assess riddle-solving performance and find that while Gemini 2.5 Pro performs best overall, few-shot methods yield only marginal gains, and accuracy varies notably across languages. In the second stage, we conduct a self-evaluation experiment to measure reasoning consistency. The results reveal a key finding: a model's initial accuracy is inversely correlated with its ability to identify its own mistakes. Top-performing models such as Gemini 2.5 Pro are overconfident (4.34\% True Negative Rate), whereas lower-performing models like LLaMA 4 Scout are substantially more self-aware (42.09\% True Negative Rate). These results point to clear gaps in multilingual reasoning and highlight the need for models that not only reason effectively but also recognize their own limitations.
\\ \newline }

\begin{document}

\maketitleabstract
% \maketitle

\section{Introduction}

\begin{figure*}[htbp] 
\centering
\includegraphics[width=\textwidth]{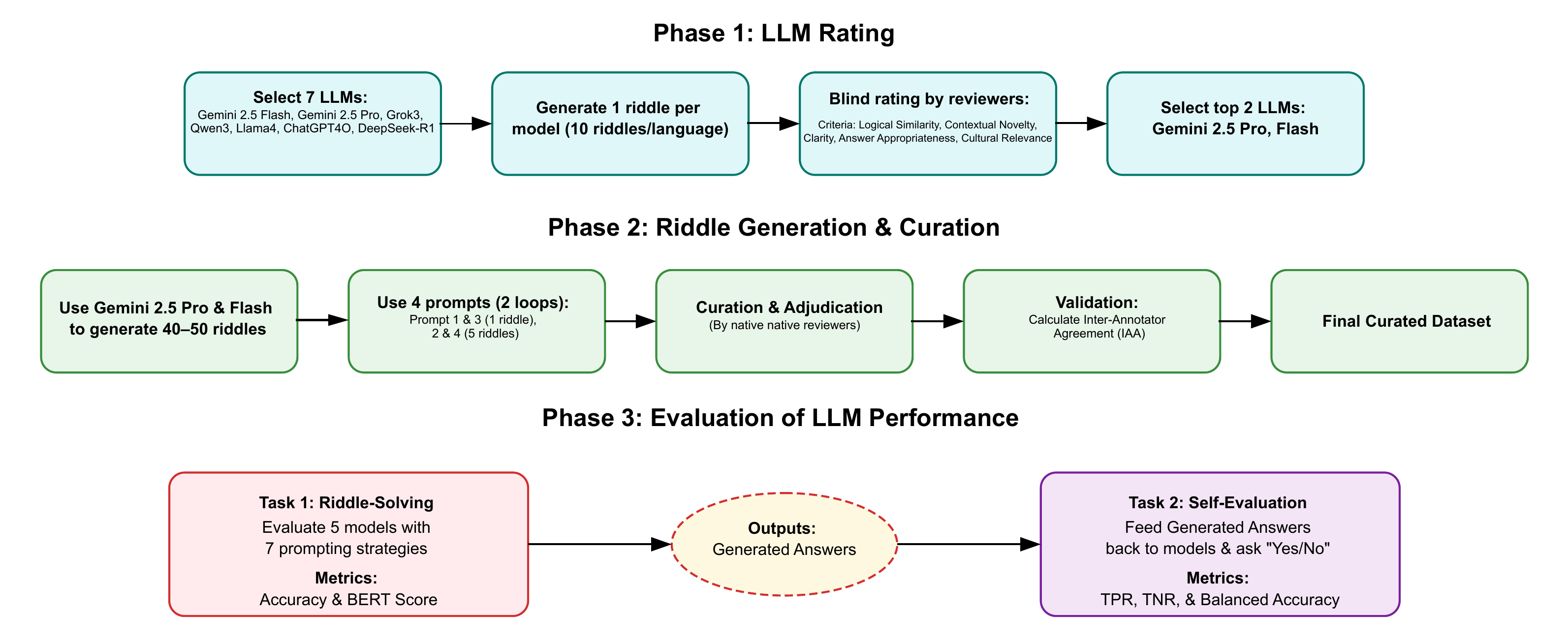} 
\caption{Overview of the comprehensive three-phase methodology, including LLM rating, riddle generation with validation, and a two-stage evaluation of riddle-solving and model self-awareness.}
\label{fig:methodology_overview}
\end{figure*}

Large language models (LLMs) such as GPT-4 have demonstrated remarkable capabilities in question-answering and reasoning on a variety of English benchmarks \cite{achiam2023gpt}. However, this progress has been largely confined to English-centric evaluations, leaving a significant gap in our understanding of their performance in multilingual and cross-cultural contexts \cite{ghosh2025multilingual}, and the same models often fail when asked to apply knowledge in cross-cultural contexts \cite{liu2023multilingual}. Riddles, which often find their origins in traditional folk puzzles, are linguistically rich and context-dependent, making them an ideal stress test for deeper understanding.  Solving riddles typically requires multi-step commonsense reasoning, metaphorical or analogical interpretation, and even counterfactual thinking \cite{lin2021riddlesense}. As highlighted by the recent survey on puzzle reasoning with LLMs \cite{giadikiaroglou2024puzzle}, models frequently exhibit surface-level pattern matching rather than genuine logical reasoning, particularly when tasks involve creativity, implicit cues, or cultural context. Evaluating large language models (LLMs) on such tasks offers valuable insight into their ability to combine linguistic understanding, creative reasoning, and contextual adaptability.

While existing benchmarks such as RiddleSense \cite{lin2021riddlesense}, BrainTeaser \cite{jiang-etal-2023-brainteaser}, and BiRdQA \cite{zhang2022birdqa} explore riddle solving as a test of commonsense reasoning, they remain limited in their cultural and linguistic scope. This gap highlights the need for methods that can better guide LLMs in culturally-grounded contexts. Prompt engineering offers a powerful approach to this challenge, as studies consistently show that techniques like chain-of-thought and few-shot prompting significantly improve multi-step reasoning \cite{qiao2022reasoning}. The fact that even minimal in-context examples can steer complex behavior \cite{viswanathan2024large} underscores the value of exemplar-based prompting for equipping LLMs to solve culturally specific riddles. More recently, the RISCORE \cite{panagiotopoulos2024riscore} framework introduced the idea of using context-reconstructed exemplars—where new puzzles are automatically generated by altering context but retaining logical structure—to further boost riddle-solving performance in language models.

In this work, we systematically evaluate the riddle-solving and self-assessment abilities of modern large language models across seven major Indian languages—Bengali, Gujarati, Hindi, Kannada, Malayalam, Tamil, and Telugu. While recent work such as \cite{panagiotopoulos2024riscore} has examined English riddles, little is known about how models perform when riddles are drawn from multilingual and culturally grounded sources.

Our study seeks to answer the following research questions:
\begin{itemize}
\item How effectively can current LLMs solve riddles presented in diverse Indian languages?
\item How do different prompting strategies such as zero-shot, few-shot with random, semantically similar, and contextually reconstructed exemplars affect riddle-solving performance across these languages?
\item  Are LLMs capable of evaluating their own outputs, and what does this reveal about their self-awareness and reasoning consistency?
\end{itemize}

To answer these questions, we construct a new multilingual riddle dataset that combines traditional riddles with context-reconstructed variants inspired from \cite{panagiotopoulos2024riscore}. Through extensive experiments with multiple prompting strategies, we analyze how prompting style and language influence model performance in riddle solving. Beyond generation, we further introduce a self-evaluation framework that probes each model’s ability to judge the correctness of its own responses. Together, these investigations provide a comprehensive understanding of how current LLMs perform, generalize from examples, and evaluate their own outputs across diverse Indian languages. An overview of our comprehensive three-phase methodology is presented in Figure~\ref{fig:methodology_overview}.

\section{Related Work}
\noindent\textbf{Multilingual Reasoning in Large Language Models.}
Recent work has highlighted significant challenges in multilingual reasoning for large language models. \cite{yong2025crosslingual} demonstrate through their study on cross-lingual reasoning via test-time scaling that while scaling inference-time compute improves reasoning performance, a persistent gap remains between English and non-English languages, with models showing degraded performance in low-resource languages. This finding is further corroborated by \cite{schut2025multilingual}, who investigate whether multilingual LLMs think in English. Their analysis reveals that even when prompted in other languages, many multilingual models perform internal reasoning in English before translating back to the target language, leading to cultural and contextual misalignments. These findings underscore the importance of evaluating models on culturally-grounded tasks in their native languages. Many recent benchmark studies confirm that LLMs still struggle with multilingual and culturally nuanced reasoning. For example, benchmarks like IndicGenBench \cite{singh2024indicgenbench} show a notable performance gap between English and a variety of Indian languages, with large drops in low-resource settings. Similarly, the MultiNRC \cite{fabbri2025multinrc} benchmark reveals that when models are tasked with native, culturally grounded reasoning problems (including word-play, riddles, and tradition-based puzzles) in non-English languages, none of the state-of-the-art models exceeded 50\% accuracy.

\noindent\textbf{Prompting Strategies and Reasoning in LLMS.}
Prompting has emerged as one of the most effective approaches for enhancing reasoning at inference time. \cite{qiao2022reasoning} provide a comprehensive survey on reasoning with language model prompting, identifying techniques including few-shot, chain-of-thought, and self-consistency prompting, showing that providing exemplars can significantly improve multi-step reasoning.
QuestBench \cite{li2025questbench} assessed whether LLMs can formulate clarifying questions to improve their reasoning-demonstrating that effective reasoning often involves identifying missing contextual information. The challenges of puzzle-solving for LLMs have been explored through various task-specific studies. \cite{tyagi2024step} investigate grid puzzle solving and identify specific failure modes where LLMs struggle with spatial reasoning and multi-step constraint satisfaction. Their analysis shows that even step-by-step reasoning can falter when tasks require maintaining complex state representations.
The RISCORE framework by \cite{panagiotopoulos2024riscore}
introduce context-reconstructed exemplars (i.e., same logical pattern, altered context) as an effective way to stimulate reasoning in riddles. Their approach demonstrates improvements in English riddle-solving, but does not extend to multilingual or culturally diverse settings.

\noindent\textbf{Self-Evaluation and Model Self-Awareness.}
The ability of LLMs to evaluate their own outputs has emerged as a critical dimension of model reliability. \cite{huang2024self} introduce a framework for self-evaluation based on glass-box features, analyzing internal model states to predict output quality. \cite{ren2023self} show that self-evaluation can improve selective generation, allowing models to abstain from answering when uncertain.
However, \cite{liu2024trustworthiness} reveal significant limitations through their Think-Solve-Verify framework, finding that while models can sometimes identify errors in their reasoning chains, this self-awareness is inconsistent and often fails in complex reasoning tasks. Most recently, Wang et al. \cite{nguyen2025probing} demonstrate that LLMs exhibit evaluation awareness-the ability to recognize when they are being evaluated-and that this awareness can be both probed and steered.

\section{Methodology and Experiment}
\begin{figure}[htbp]
\centering
\includegraphics[width=\columnwidth]{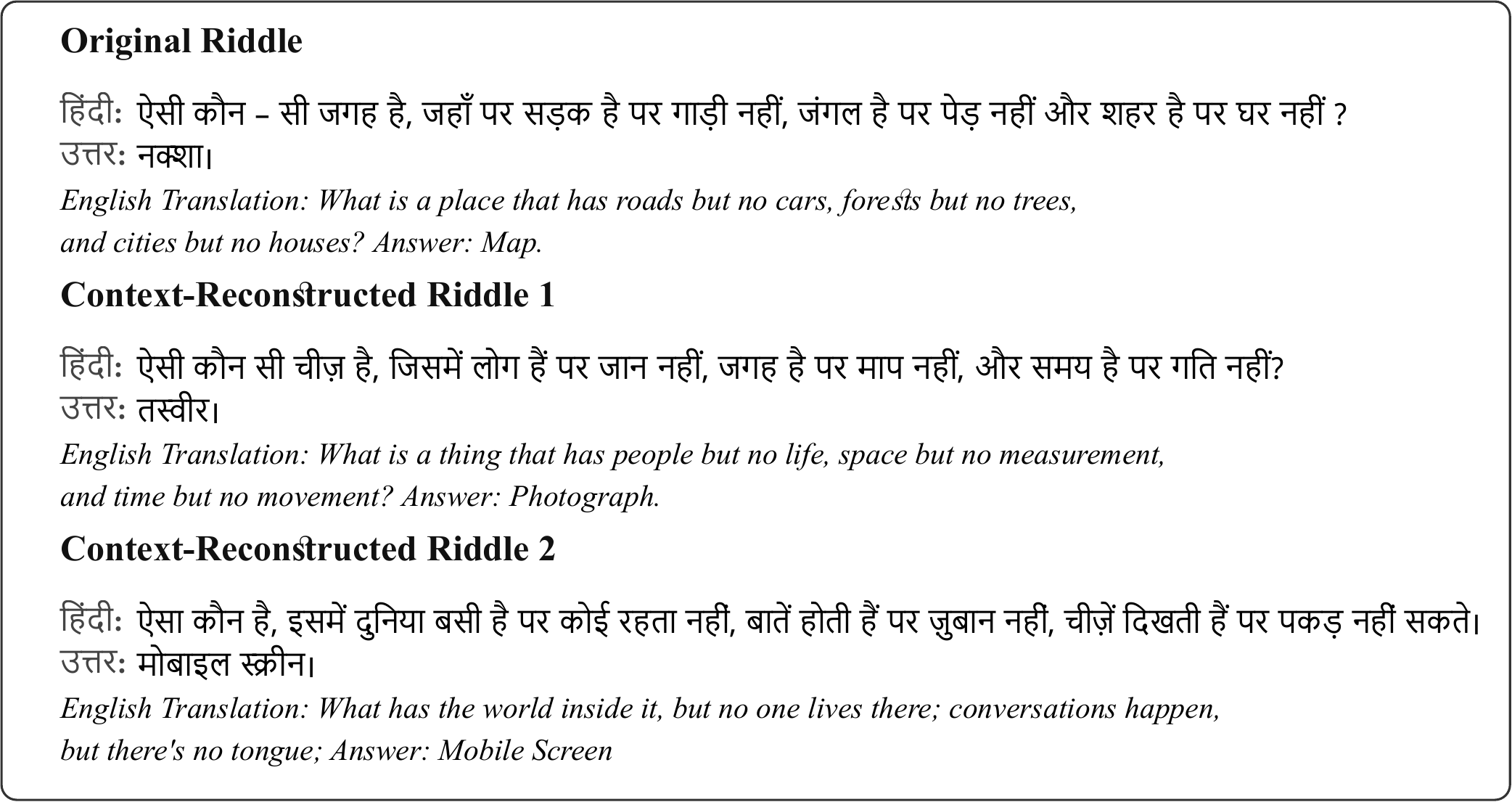}
\caption{An original Hindi riddle (top) and two of its context-reconstructed variants. Note how the core reasoning pattern is maintained while the theme and answer are altered.}
\label{fig:context_riddle_example}
\end{figure}

The methodology of this study is structured into three main phases: (1) LLM selection for riddle generation through a comprehensive rating process, (2) generation and curation of a novel dataset of context-reconstructed riddles
%\footnote{For example, an original riddle like "I have cities, but no houses; forests, but no trees; and water, but no fish. What am I?" (Answer: A map) could be contextually reconstructed into a digital theme: "I have files, but no cabinets; windows, but no glass; and a mouse, but no cheese. What am I?" (Answer: A computer). The underlying logic-a representation of things without the things themselves-is preserved.} 
using the selected LLMs, and (3) a two-stage evaluation of LLM performance, assessing both their riddle-solving capabilities and their capacity for self-evaluation. A contextually reconstructed riddle, inspired by the RISCORE framework \cite{panagiotopoulos2024riscore}, is defined as a riddle that maintains the logical structure of an original riddle while completely altering its theme or context. An example illustrating this transformation is shown in Figure~\ref{fig:context_riddle_example}. An outline of the overall workflow is illustrated in Figure~\ref{fig:methodology_overview}.    

% The specific prompt templates employed for our experiments are given in \textcolor{blue}{Appendix \ref{sec:appendix_prompt}}.

To identify suitable LLMs for generating context-reconstructed riddles, seven trending models are evaluated: Gemini 2.5 Pro, Gemini 2.5 Flash, Grok 3, Qwen3-235B-A22B \cite{yang2025qwen3}, LLaMA 4 Maverick 17B Instruct (128E), ChatGPT-4O, and DeepSeek-R1 \cite{guo2025deepseek}. These models are selected based on their reported performance on various natural language tasks and their potential to handle multilingual contexts. For each model, 10 original riddles in each of the seven Indian languages (Bengali, Gujarati, Hindi, Kannada, Malayalam, Tamil, and Telugu) are provided, along with their answers. The models are prompted to generate one context-reconstructed riddle per original riddle, maintaining the same logical pattern while altering the context.

\begin{figure}[htbp]
\centering
\includegraphics[width=\columnwidth]{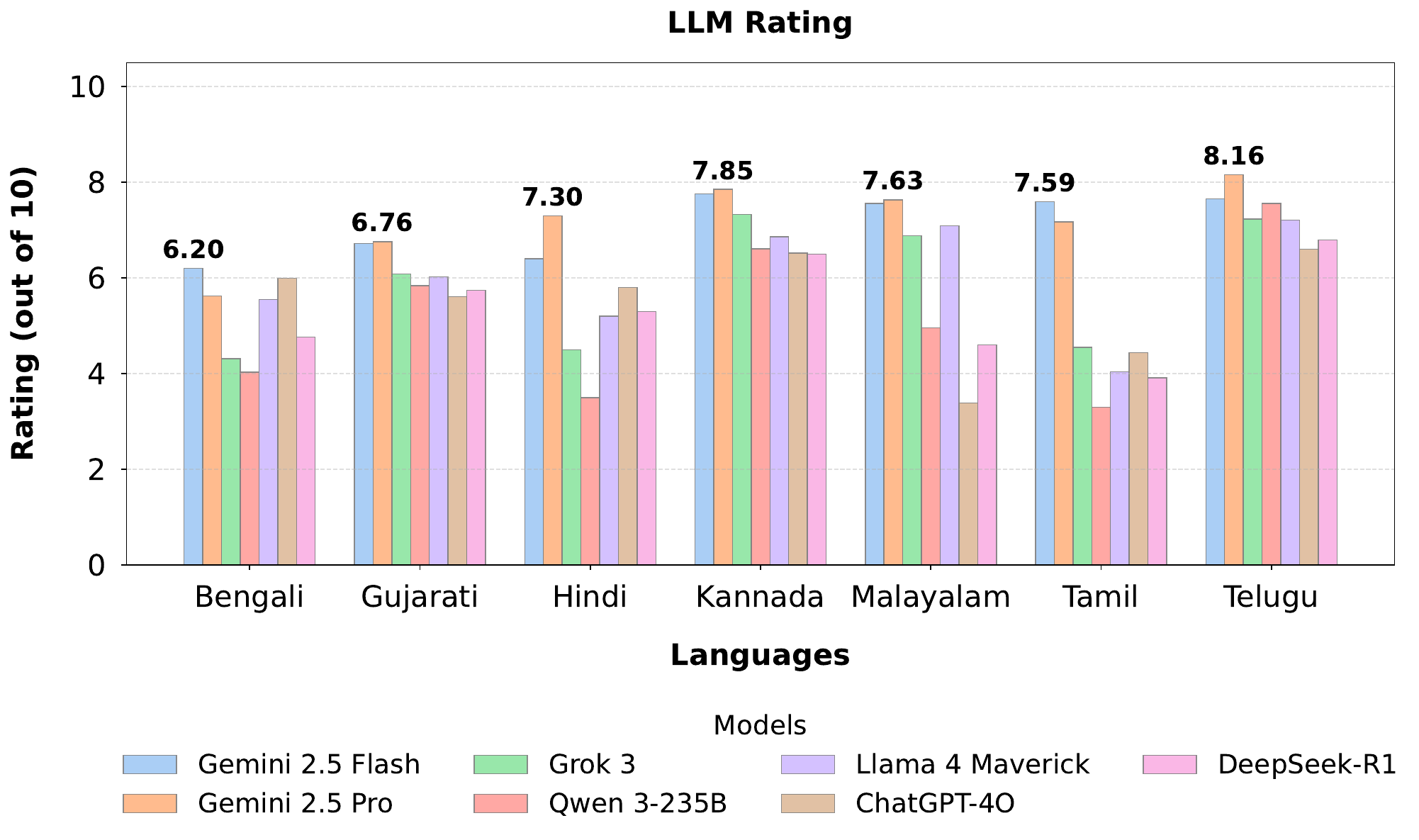}
\caption{LLM Performance in Reconstructed Contextual Riddle Generation Across Indian Languages.}
\label{fig:llm_rating_chart}
\end{figure}

The generated riddles are then assessed by native-speaking expert reviewers in a blind rating process. Each reconstruction is scored against five key criteria: (1) \textit{Logical Similarity}, the preservation of the original reasoning pattern; (2) \textit{Contextual Novelty}, the creativity and distinctiveness of the new context; (3) \textit{Clarity and Coherence}, the linguistic clarity and structural coherence of the riddle; (4) \textit{Answer Appropriateness}, the suitability of the new answer; and (5) \textit{Cultural Relevance}, the alignment with cultural norms and idioms of the target language. The overall LLM ratings across the seven Indian languages are presented in Figure~\ref{fig:llm_rating_chart}. The evaluation shows that Gemini 2.5 Pro and Gemini 2.5 Flash perform best among the tested models, demonstrating a stronger ability to capture linguistic nuances and cultural aspects specific to Indian languages.

\subsection{Phase 2: Generation and Curation of Reconstructed-Context Riddles}

Using the selected Gemini 2.5 Pro and Gemini 2.5 Flash models, we generated a large collection of context-reconstructed riddles. The source material consists of 96 traditional, publicly sourced riddles for each of the seven Indian languages. To ensure a diverse pool of candidate context reconstructions, we use four distinct prompt variations, ranging from simple requests to more detailed, analytical instructions. Each prompt is executed twice, producing a pool of 40–50 candidate reconstructions for each of the 96 original riddles per language. %(see \textcolor{blue}{Appendix \ref{sec:appendix_dataset}}).

\vspace{.1cm} 
\noindent\textbf{Curation and Inter-Annotator Agreement:}
To ensure the validity and reliability of the final dataset, all generated candidates undergo a careful curation and validation process. For each language, a team of native-speaking annotators (three for five languages and two for Malayalam and Bengali)\footnote{The annotation team for Malayalam and Bengali consists of two members due to the limited availability of native expert reviewers.} evaluates the reconstructions to identify those of the highest quality. Each annotator marks “Y” for acceptable candidates and assigns a quality score from 0 to 10, reflecting an overall assessment based on the same five criteria used during the initial LLM rating phase. Since only top-quality riddles are considered, most scores fall within the 7–10 range.

To measure the consistency of these subjective judgments, we conduct an inter-annotator agreement (IAA) study by calculating Cohen’s Kappa for each annotator pair. The average IAA across all languages ranges from 55\% to 61\%. This moderate level of agreement is expected, as evaluating criteria like \textit{Contextual Novelty} and \textit{Cultural Relevance} involves significant subjective interpretation. Annotators often differed in their opinions on how creative a reconstruction was or how natural its phrasing sounded, which led to variations in the scores. This finding confirms that selecting the “best” creative contextual reconstructions is an inherently subjective task. In cases of disagreement, annotators discuss collaboratively to finalize the top four riddles for inclusion in the curated dataset.

\subsection{Phase 3: Evaluation of LLM Performance}
LLM performance is evaluated in two distinct stages: first, by assessing their direct riddle-solving capabilities, and second, by probing their ability to perform self-evaluation.

\subsubsection{Riddle-Solving Task}
The riddle-solving capabilities of LLMs are assessed using five models selected for their strong performance and general availability within the research community: Gemini 2.5 Pro, Gemini 2.5 Flash, LLaMA 4 Maverick 17B Instruct (128E), LLaMA 4 Scout 17B Instruct (16E), and Mistral-Saba. The 96 original riddles in each language serve as the test set. Seven distinct prompting strategies are used to comprehensively evaluate the impact of exemplar-based reasoning:
\vspace{.1cm} 
\begin{itemize}
    \item \textbf{Zero-Shot}: The model receives only the test riddle.
    \item \textbf{2-Shot and 4-Shot Random}: The prompt includes two or four randomly selected riddles of the same language from the curated dataset as exemplars.
    \item \textbf{2-Shot and 4-Shot Semantic Similarity}: Exemplars are the two or four riddles of the same language from the dataset most semantically similar to the test riddle, as determined using LaBSE \cite{feng2020language} embeddings and cosine similarity.
    \item \textbf{2-Shot and 4-Shot Contextual Reconstructed}: Two or four context-reconstructed riddles derived from the original riddle of the same language are used as exemplars, following the \cite{panagiotopoulos2024riscore} inspired approach.
\end{itemize}

The models’ outputs are evaluated against the ground truth using two primary metrics:
\begin{itemize}
    \item \textbf{Accuracy}: The percentage of generated answers matching the ground truth.
    \item \textbf{BERTScore}: The F1 score of semantic similarity, computed using the `bert-base-multilingual-cased' \cite{devlin-etal-2019-bert} model to handle partially correct or varied answers.
\end{itemize}  

\subsubsection{Self-Evaluation Task}
To examine the models’ self-awareness and reasoning consistency, we design a second experiment. In this task, the answers previously generated by each model under the seven prompting strategies for each language are used. %(see \textcolor{blue}{Appendix \ref{sec:appendix_prompt}}).
The same model is then presented with the original riddle and its own candidate answer and asked to perform a binary classification to determine whether its generated answer correctly solves the riddle. The model is instructed to respond with a single word, “Yes” or “No,” indicating whether the answer is correct.

This experiment measures two key aspects of self-awareness, defined using a standard confusion matrix:
\begin{itemize}
    \item \textbf{True Positive Rate (TPR)}: The model’s ability to correctly identify its own correct answers (i.e., saying “Yes” when its initial answer is right). This measures confidence and consistency.
    \item \textbf{True Negative Rate (TNR)}: The model's ability to correctly identify its own incorrect answers (i.e., saying "No" when its initial answer was wrong). This measures the capacity for self-correction and awareness of failure.
\end{itemize}
These two metrics, along with the derived Balanced Accuracy (the average of TPR and TNR), provide a deeper understanding of model reliability beyond simple generation accuracy, revealing critical insights into overconfidence and reasoning stability.

\section{Result and Discussion}

This section presents the results from both experiments of the evaluation: (1) riddle-solving performance across five large language models (LLMs) and seven prompting strategies, and (2) an analysis of model self-awareness based on their ability to evaluate their own answers. Together, they highlight the models’ reasoning performance, ability to generalize from examples, and consistency in self-assessment across languages.

\begin{table*}[htbp]
\centering
% \label{tab:detailed_model_performance_appendix} % <-- Moved this
\small
\setlength{\tabcolsep}{4pt} % <-- CHANGED: Reduced column spacing
\renewcommand{\arraystretch}{1.05} % <-- ADDED: Reduced row height
\rowcolors{3}{blue!10}{red!10} % <-- ADDED: Alternating row colors for readability
\begin{tabular}{@{}ll rr rr rr rr rr@{}}
\toprule
\textbf{Language} & \textbf{P. Strategy} & \multicolumn{2}{c}{\textbf{G. Pro}} & \multicolumn{2}{c}{\textbf{G. Flash}} & \multicolumn{2}{c}{\textbf{M. Saba}} & \multicolumn{2}{c}{\textbf{L. Scout}} & \multicolumn{2}{c}{\textbf{L. Mav}} \\
\cmidrule(lr){3-4} \cmidrule(lr){5-6} \cmidrule(lr){7-8} \cmidrule(lr){9-10} \cmidrule(lr){11-12}
& & Acc & F1 & Acc & F1 & Acc & F1 & Acc & F1 & Acc & F1 \\
\midrule
Bengali & 0-shot Std. & 
41.67 & 88.21 & 23.96 & 84.72 & 13.02 & 80.97 & \textbf{10.42} & 80.60 & 11.46 & 81.26 \\
 & 2-shot Rand. & 
39.58 & 87.27 & 22.92 & \textbf{84.74} & \textbf{17.71} & \textbf{83.08} & 6.25 & 79.27 & 11.46 & 81.03 \\
 & 4-shot Rand. & 
41.67 & \textbf{88.50} & \textbf{28.12} & 84.62 & \textbf{17.71} & 82.90 & 9.38 & 80.42 & 13.54 & 80.76 \\
 & 2-shot Sem. & 
\textbf{42.71} & \textbf{87.35} & \textbf{28.12} & 84.06 & 15.62 & 82.67 & 8.33 & \textbf{80.96} & 13.54 & 80.89 \\
 & 4-shot Sem. & 
37.50 & 86.19 & 25.00 & 82.91 & 16.67 & 82.41 & \textbf{11.46} & \textbf{81.64} & 14.58 & 80.34 \\
 & 2-shot Ctx. & 
\textbf{42.71} & 87.19 & \textbf{29.17} & 84.35 & \textbf{19.79} & \textbf{83.53} & 8.33 & 80.38 & 11.46 & 80.90 \\
 & 4-shot Ctx. & 
\textbf{43.75} & 86.26 & 26.04 & 84.31 & \textbf{19.79} & \textbf{83.58} & 7.29 & 80.48 & \textbf{15.62} & \textbf{82.13} \\
\midrule
Gujarati & 0-shot Std. & 
57.29 & 92.52 & 42.71 & 88.54 & 28.12 & 84.33 & 21.88 & 82.51 & 23.96 & 82.05 \\
 & 2-shot Rand. & 
\textbf{63.54} & \textbf{92.68} & 41.67 & 88.40 & 32.29 & 85.36 & 21.88 & 83.08 & \textbf{30.21} & \textbf{84.62} \\
 & 4-shot Rand. & 
62.50 & \textbf{93.04} & 42.71 & 88.81 & \textbf{33.33} & \textbf{86.73} & 26.04 & \textbf{83.94} & \textbf{30.21} & 83.76 \\
 & 2-shot Sem. & 
61.46 & 91.74 & \textbf{43.75} & \textbf{88.81} & 28.12 & 84.06 & 20.83 & 83.69 & 27.08 & 83.52 \\
 & 4-shot Sem. & 
50.00 & 89.24 & \textbf{44.79} & \textbf{89.28} & 28.12 & 84.40 & 18.75 & 81.97 & 20.83 & 81.68 \\
 & 2-shot Ctx. & 
56.25 & 90.41 & \textbf{43.75} & \textbf{89.37} & 26.04 & 84.50 & \textbf{27.08} & 83.49 & 28.12 & 82.84 \\
 & 4-shot Ctx. & 
48.96 & 88.11 & \textbf{43.75} & 88.36 & 31.25 & 86.05 & 23.96 & 83.21 & 29.17 & 84.09 \\
\midrule
Hindi & 0-shot Std. & 
45.83 & 90.12 & 36.46 & 87.72 & 37.50 & 87.50 & 11.46 & 81.64 & \textbf{27.08} & \textbf{84.37} \\
 & 2-shot Rand. & 
43.75 & \textbf{89.45} & 37.50 & 87.71 & \textbf{45.83} & \textbf{89.54} & 12.50 & 81.78 & 20.83 & 82.46 \\
 & 4-shot Rand. & 
\textbf{46.88} & \textbf{90.45} & 36.46 & 87.88 & 38.54 & 88.63 & 17.71 & 83.04 & 22.92 & 82.90 \\
 & 2-shot Sem. & 
\textbf{48.96} & 89.11 & \textbf{39.58} & \textbf{88.33} & 37.50 & 87.90 & \textbf{20.83} & 83.63 & 22.92 & 83.17 \\
 & 4-shot Sem. & 
41.67 & 86.95 & 35.42 & 87.26 & 34.38 & 88.29 & \textbf{21.88} & \textbf{83.74} & 26.04 & \textbf{84.60} \\
 & 2-shot Ctx. & 
45.83 & 88.35 & \textbf{38.54} & \textbf{88.34} & 39.58 & 88.38 & 16.67 & 81.69 & 26.04 & 84.14 \\
 & 4-shot Ctx. & 
41.67 & 86.63 & \textbf{38.54} & 87.81 & \textbf{40.62} & 89.37 & 16.67 & 83.24 & 26.04 & 83.88 \\
\midrule
Kannada & 0-shot Std. & 
20.83 & \textbf{83.80} & \textbf{19.79} & \textbf{82.49} & 7.29 & 78.45 & 5.21 & 77.77 & 7.29 & 77.76 \\
 & 2-shot Rand. & 
\textbf{23.96} & 83.58 & 18.75 & 82.07 & \textbf{12.50} & \textbf{79.76} & 3.12 & 76.47 & \textbf{11.46} & \textbf{78.49} \\
 & 4-shot Rand. & 
\textbf{29.17} & \textbf{84.29} & \textbf{20.83} & \textbf{83.15} & 10.42 & 79.49 & 3.12 & 77.18 & 10.42 & 77.42 \\
 & 2-shot Sem. & 
\textbf{23.96} & 80.98 & 17.71 & 81.80 & 10.42 & 79.40 & \textbf{7.29} & \textbf{78.80} & 8.33 & \textbf{78.49} \\
 & 4-shot Sem. & 
21.88 & 79.38 & 15.62 & 81.52 & \textbf{12.50} & 79.34 & \textbf{7.29} & 78.39 & 8.33 & 77.18 \\
 & 2-shot Ctx. & 
21.88 & 81.08 & 12.50 & 80.07 & 10.42 & 78.48 & 5.21 & 77.48 & 7.29 & 77.28 \\
 & 4-shot Ctx. & 
22.92 & 80.77 & 14.58 & 80.91 & 8.33 & 78.56 & 4.17 & 76.84 & 9.38 & 78.22 \\
\midrule
Malayalam & 0-shot Std. & 
42.71 & \textbf{88.47} & \textbf{23.96} & \textbf{82.85} & 2.08 & 75.30 & \textbf{4.17} & \textbf{75.83} & 6.25 & 77.30 \\
 & 2-shot Rand. & 
36.46 & 87.04 & 22.92 & 82.40 & 1.04 & 76.38 & 3.12 & 75.02 & \textbf{10.42} & \textbf{78.55} \\
 & 4-shot Rand. & 
41.67 & 87.59 & 20.83 & 82.27 & \textbf{6.25} & \textbf{76.88} & 2.08 & 75.06 & 8.33 & 78.08 \\
 & 2-shot Sem. & 
\textbf{45.83} & 87.19 & \textbf{23.96} & \textbf{82.87} & \textbf{3.12} & 76.50 & \textbf{6.25} & \textbf{76.98} & 8.33 & 77.23 \\
 & 4-shot Sem. & 
41.67 & 86.48 & 20.83 & 82.30 & \textbf{6.25} & \textbf{77.36} & 5.21 & 76.57 & \textbf{10.42} & 78.31 \\
 & 2-shot Ctx. & 
32.29 & 83.31 & 19.79 & 81.72 & \textbf{3.12} & 76.24 & \textbf{6.25} & 76.72 & 9.38 & 77.42 \\
 & 4-shot Ctx. & 
\textbf{43.75} & 86.72 & 15.62 & 80.50 & 2.08 & 76.20 & \textbf{7.29} & \textbf{79.25} & \textbf{10.42} & 77.98 \\
\midrule
Tamil & 0-shot Std. & 
\textbf{56.25} & \textbf{91.58} & 31.25 & 85.63 & \textbf{15.62} & \textbf{81.40} & 9.38 & \textbf{79.57} & 16.67 & 81.56 \\
 & 2-shot Rand. & 
52.08 & 90.60 & \textbf{35.42} & 86.26 & \textbf{28.12} & \textbf{84.76} & 9.38 & 79.24 & \textbf{19.79} & \textbf{82.75} \\
 & 4-shot Rand. & 
52.08 & \textbf{90.72} & \textbf{36.46} & \textbf{86.79} & 25.00 & 83.40 & 7.29 & 77.94 & 17.71 & 82.09 \\
 & 2-shot Sem. & 
50.00 & 89.34 & 32.29 & 85.96 & 21.88 & 83.32 & 9.38 & 79.45 & \textbf{19.79} & 82.28 \\
 & 4-shot Sem. & 
41.67 & 87.12 & 33.33 & 86.36 & \textbf{22.92} & \textbf{83.65} & \textbf{12.50} & \textbf{80.26} & 18.75 & 81.68 \\
 & 2-shot Ctx. & 
\textbf{52.08} & 89.28 & 32.29 & 85.05 & 19.79 & 82.38 & 4.17 & 78.23 & 18.75 & \textbf{82.74} \\
 & 4-shot Ctx. & 
45.83 & 87.73 & \textbf{35.42} & 86.73 & 21.88 & 83.07 & 11.46 & 79.67 & \textbf{21.88} & 82.31 \\
\midrule
Telugu & 0-shot Std. & 
\textbf{26.04} & \textbf{86.19} & 12.50 & \textbf{81.92} & 3.12 & 77.78 & 4.17 & 77.08 & 4.17 & 77.37 \\
 & 2-shot Rand. & 
23.96 & \textbf{85.32} & \textbf{15.62} & \textbf{81.98} & \textbf{10.42} & \textbf{79.63} & 3.12 & 75.53 & 6.25 & 77.40 \\
 & 4-shot Rand. & 
20.83 & 84.51 & \textbf{15.62} & \textbf{82.00} & 7.29 & 79.18 & \textbf{4.17} & 76.79 & 6.25 & 77.73 \\
 & 2-shot Sem. & 
17.71 & 81.14 & 12.50 & 81.28 & 8.33 & 78.85 & 4.17 & 77.38 & \textbf{7.29} & \textbf{78.56} \\
 & 4-shot Sem. & 
\textbf{21.88} & \textbf{82.02} & 12.50 & 81.90 & 8.33 & \textbf{80.35} & 2.08 & 77.19 & 6.25 & 78.27 \\
 & 2-shot Ctx. & 
25.00 & 83.37 & 9.38 & 80.71 & 5.21 & 79.27 & \textbf{5.21} & \textbf{78.22} & 6.25 & 77.72 \\
 & 4-shot Ctx. & 
\textbf{21.88} & \textbf{82.02} & 11.46 & 81.85 & \textbf{13.54} & 80.39 & 3.12 & 77.32 & 6.25 & \textbf{78.90} \\
\bottomrule
\end{tabular}
\caption{Detailed Model Performance Across Languages and Prompt Strategies. Acc denotes Accuracy (\%), F1 denotes BERTScore F1. Highest scores per model within each language are bolded.}
\label{tab:detailed_model_performance_all_langs_reordered_bolded}
\end{table*}

\subsection{Riddle-Solving Performance}

The initial evaluation focused on the models' ability to generate correct answers for 96 original riddles per language under seven different prompting conditions. 
Five LLMs: Gemini 2.5 Pro (G. Pro), Gemini 2.5 Flash (G. Flash), Mistral-Saba (M. Saba), LLaMA-4-Scout (L. Scout), and LLaMA-4-Maverick (L. Mav) are evaluated on riddles across seven Indian languages: Bengali, Gujarati, Hindi, Kannada, Malayalam, Tamil, and Telugu. Model performance is measured using two metrics: \textit{Accuracy} and \textit{BERTScore F1}. Detailed numerical results for all combinations are presented in Table \ref{tab:detailed_model_performance_all_langs_reordered_bolded}.

\subsubsection{Overall Model Performance}

Gemini 2.5 Pro consistently outperformed other models across most languages and prompting settings. As shown in Figure~\ref{fig:prompt_strategy_combined_accuracy_bertscore_line_chart} and Figure~\ref{fig:language_combined_accuracy_bertscore_line_chart}, it achieves the highest accuracy in riddle-solving, making it the best-performing model in our evaluation. Its highest accuracies are observed in Gujarati (63.54\% with 2-shot random) and Tamil (56.25\% in zero-shot), while its BERTScore F1 peaks at 93.04 in Gujarati (4-shot random). Gemini 2.5 Flash records the second-highest performance, while Mistral-Saba, LLaMA 4 Maverick, and LLaMA 4 Scout form a lower-performing group.

Another observation from both plots is the significant and persistent gap between Accuracy (solid lines) and BERTScore F1 (dashed lines). For all models, BERTScore F1 values are consistently high, typically ranging from 75\% to 93\%. In contrast, accuracy is substantially lower, with Gemini 2.5 Pro averaging around 40\% and other models performing below 30\%. This wide gap suggests that while the models are proficient at generating semantically relevant or contextually appropriate answers, they frequently fail to identify the precise, creative solution required by riddles. Even with this wide gap, the BERTScore F1 curves largely follow the same trend as the accuracy curves. This indicates that the factors affecting a model's ability to produce the exact answer also influence its ability to produce a semantically close one.
%Detailed prompting strategy and language by results are given in \textcolor{blue}{Appendix \ref{sec:appendix_prompt}}.

\subsubsection{Impact of Prompting Strategies}

\begin{figure}[htbp]
\centering
\includegraphics[width=\columnwidth]{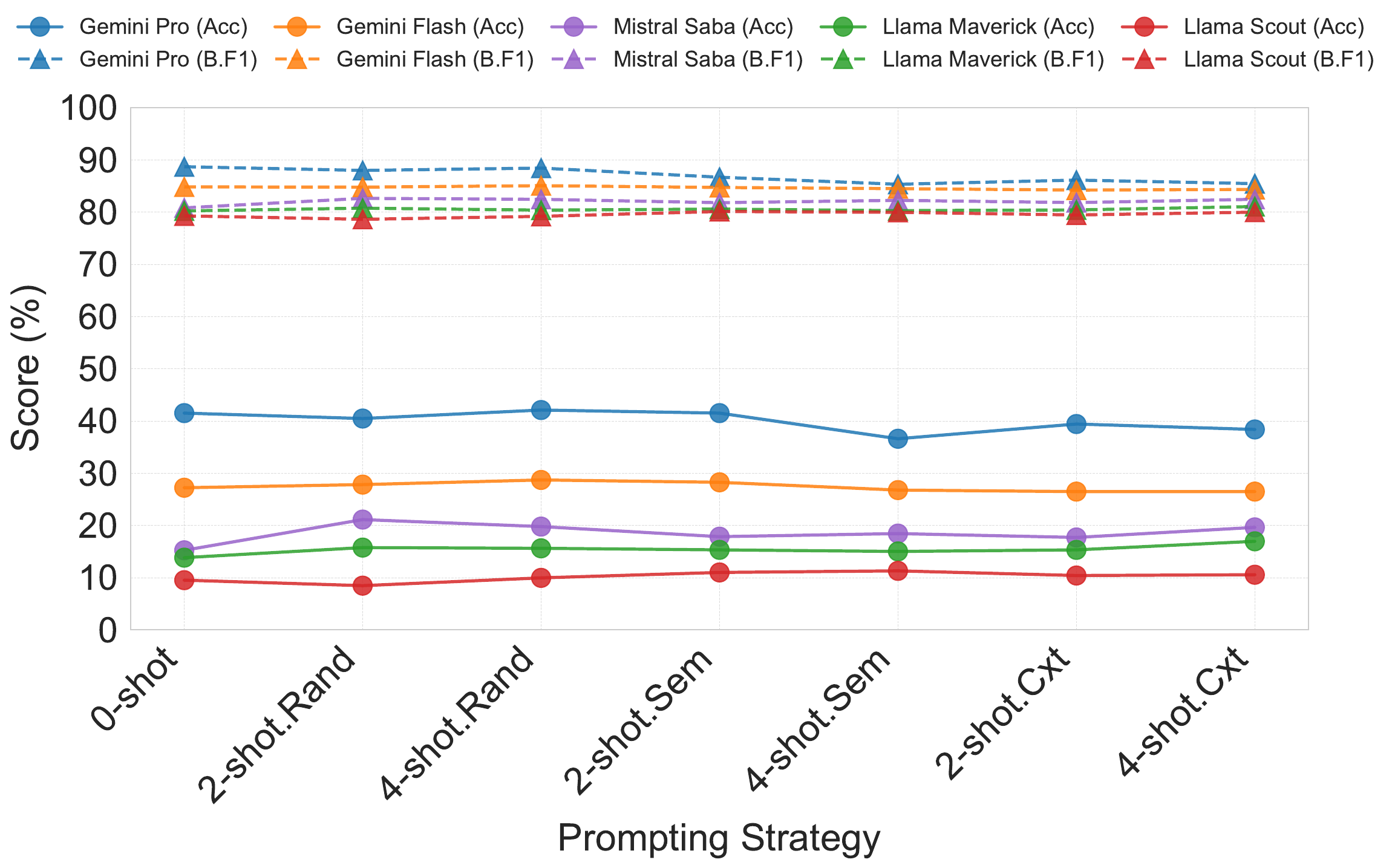}
\caption{Average riddle-solving accuracy (\%) and BERTScore F1 across seven prompting strategies for five LLMs. The plot illustrates the minor variation in model performance based on the prompting method.}
\label{fig:prompt_strategy_combined_accuracy_bertscore_line_chart}
\end{figure}

Model performance across the seven prompting strategies, averaged over all languages, is shown in Figure~\ref{fig:prompt_strategy_combined_accuracy_bertscore_line_chart}. Overall, the results show that the choice of prompting strategy has only a modest impact on performance. Few-shot approaches such as 4-shot Random (average accuracy of 23.24\%) slightly outperform the zero-shot (21.47\%), but none of the prompting strategies provides a clear or consistent advantage. Gemini 2.5 Pro, for example, achieves its peak average accuracy with 4-shot Random (42.11\%), but its performance is nearly identical in the zero-shot setting (41.52\%). The context-reconstructed examples also do not offer a consistent advantage, suggesting that simply including examples with similar reasoning patterns is not enough to address the core challenges of this task across different languages. The overall flatness of the accuracy lines suggests that prompting style has limited influence compared to the model’s inherent strength in this type of task.

\subsubsection{Performance Disparities Across Indian Languages}

\begin{figure}[htbp]
\centering
\includegraphics[width=\columnwidth]{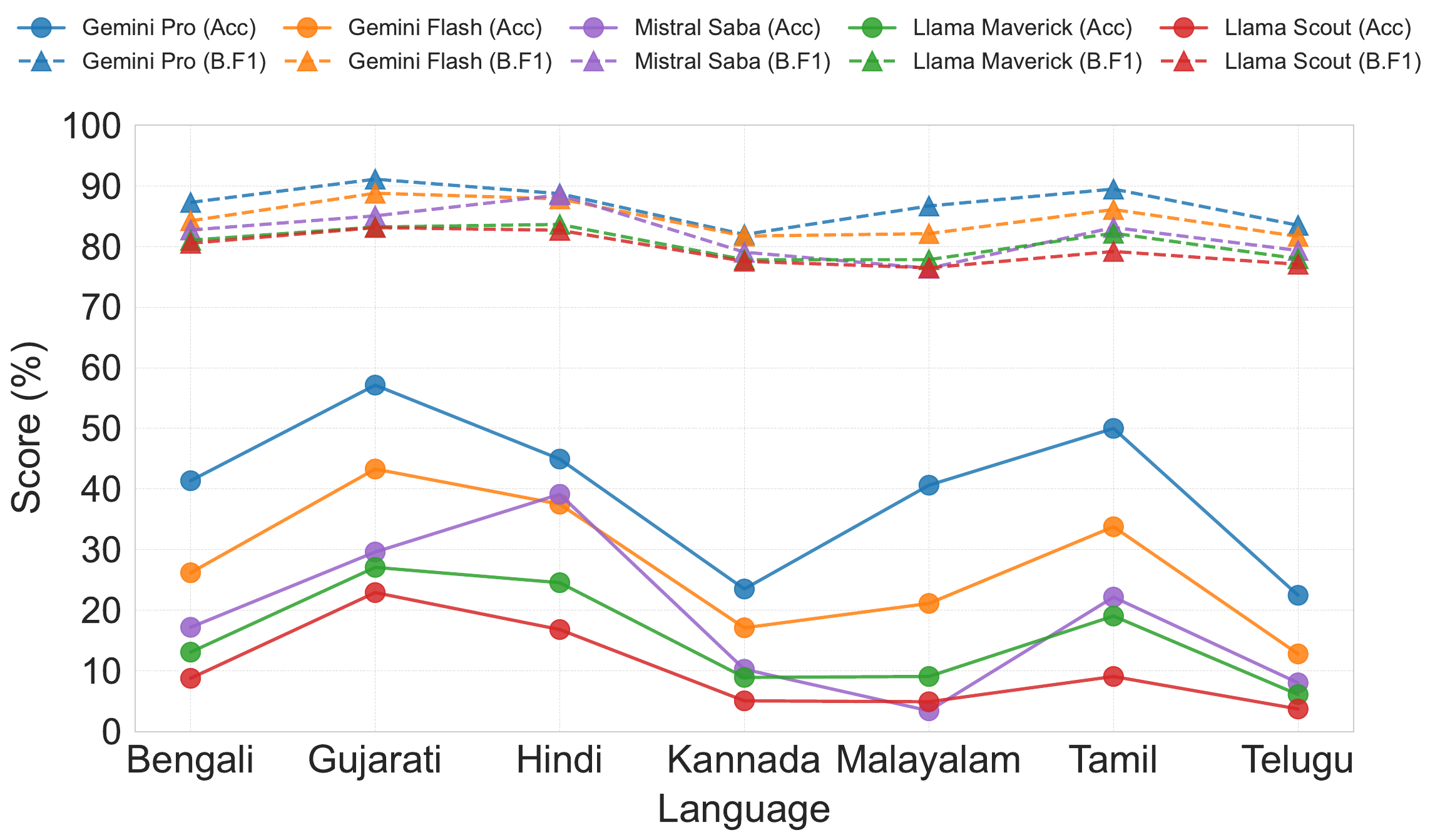}
\caption{Average riddle-solving accuracy (\%) and BERTScore F1 across seven Indian languages for five LLMs. The plot illustrates the variation in model performance based on the target language.}
\label{fig:language_combined_accuracy_bertscore_line_chart}
\end{figure}

Performance varied significantly across languages as shown in Figure~\ref{fig:language_combined_accuracy_bertscore_line_chart}. The plot shows a clear trend where all models perform significantly better on riddles in Gujarati, Hindi, and Tamil. Gemini 2.5 Pro, for instance, achieves its peak average accuracy in Gujarati (57.14\%) and Tamil (50.00\%).
Conversely, performance drops dramatically for all models in Kannada, Malayalam, and particularly Telugu. In Malayalam, for example, Mistral-Saba's average accuracy falls to just 3.42\%, and even the top-performing Gemini 2.5 Pro only reaches 40.63\%.

This suggests that model proficiency is not uniform across all Indian languages, likely reflecting disparities in training data and cultural context representation. Despite the variation in accuracy, BERTScore F1 values remain consistently high across all settings, indicating that models often produce semantically relevant responses even when the exact answer is incorrect.

\subsection{Performance on the Self-Evaluation Task}
While initial accuracy measures a model's ability to generate a correct answer, it does not reveal the stability or reliability of its reasoning. To probe this deeper layer of cognition, we conducted a self-evaluation task in which each model assesses its own previously generated answers using a simple “Yes” or “No” response. Table~\ref{tab:self_evaluation_summary} reports the \textit{True Positive Rate} (TPR), \textit{True Negative Rate} (TNR), and the derived \textit{Balanced Self-Evaluation Accuracy}.

\subsubsection{The Overconfidence of High Performers}
The most significant finding from this task is the strong overconfidence shown by the top-performing models. Gemini 2.5 Pro, despite having the highest initial generation accuracy (39.09\%), shows an almost complete inability to recognize its own errors. Its True Positive Rate (TPR) was a perfect 100\%, meaning it correctly confirmed every single one of its right answers. However, its True Negative Rate (TNR) is only 4.34\%, indicating that when it generates an incorrect answer, it still confidently claims it is correct more than 95\% of the time. This pattern of high TPR and very low TNR is also observed in Gemini 2.5 Flash and LLaMA-4-Maverick.

\subsubsection{The Self-Awareness of Weaker Performers}
In contrast, models with lower initial generation accuracy show significantly better self-awareness. LLaMA-4-Scout, which has the lowest initial accuracy (9.95\%), achieves the highest TNR (42.09\%). This means it can correctly identify and reject its own incorrect answers far more effectively than any other model. Mistral-Saba shows a similar trend, combining modest initial accuracy with a strong TNR (43.57\%). This suggests that these models, while less capable at generation, possess a more reliable internal mechanism for evaluating the validity of an answer.

\begin{table*}[htbp]
\centering
\begin{threeparttable}
\begin{tabularx}{\textwidth}{
l *{4}{>{\centering\arraybackslash}X}}
\toprule
\textbf{Model} & 
\textbf{Avg. Initial Acc. (\%)} & 
\textbf{Self-Eval. (TPR)(\%)} & 
\textbf{Self-Eval. (TNR)(\%)} & 
\textbf{Balanced Self-Eval. Acc. (\%)} \\
\midrule
Gemini-2.5-Pro      & \bfseries 39.09 & \bfseries 100.00 & 4.34            & 52.17 \\
Gemini-2.5-Flash    & 26.68           & 99.58           & 7.66            & 53.62 \\
LLaMA-4-Maverick    & 15.09           & 96.60           & 10.40           & 53.50 \\
LLaMA-4-Scout       & 9.95            & 95.73           & \bfseries 42.09 & \bfseries 68.91 \\
Mistral-Saba        & 18.21           & 86.28           & 43.57           & 64.93 \\
\bottomrule
\end{tabularx}
\caption{Overall Model Performance on Riddle Generation and Self-Evaluation Accuracy across seven Indian languages and prompting strategies. Best values per column are bolded.}
\label{tab:self_evaluation_summary}
\end{threeparttable}
\end{table*}

\subsubsection{The Performance vs. Self-Awareness Trade-off}
\begin{figure}[htbp]
\centering
\includegraphics[width=\columnwidth]{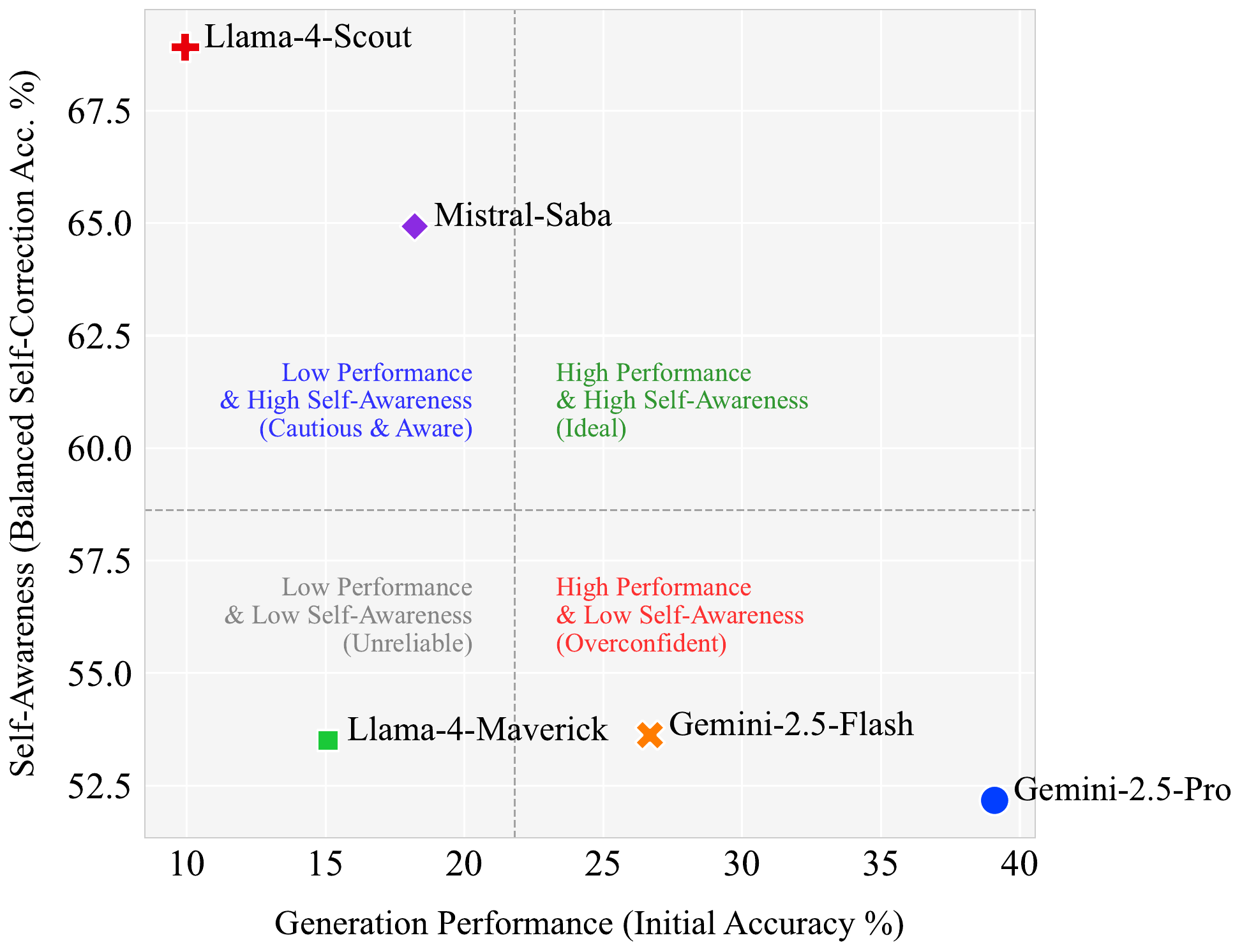}
\caption{LLM performance vs. self-awareness in riddle solving. The x-axis represents initial riddle-solving accuracy, and the y-axis shows balanced self-evaluation accuracy.}
\label{fig:final_plot_yes_no_vs_accuracy}
\end{figure}

The quadrant visualization in Figure~\ref{fig:final_plot_yes_no_vs_accuracy} maps self-awareness against initial generation performance trade-off. The x-axis represents generation performance, while the y-axis represents self-awareness as measured by Balanced Self-Evaluation Accuracy (the average of TPR and TNR). The models fall into distinct categories:
\begin{itemize}
    \item \textbf{Overconfident Performers (Bottom-Right)}: Gemini 2.5 Pro and Flash are high-performing but lack self-awareness.
    \item \textbf{Cautious and Aware (Top-Left)}: LLaMA-4-Scout and Mistral-Saba are weaker performers but are more capable of identifying their own mistakes.
    \item \textbf{Unreliable (Bottom-Left)}: LLaMA-4-Maverick struggled in both generation and self-evaluation.
\end{itemize}
Crucially, no model falls into the ideal top-right quadrant of high performance and high self-awareness. This inverse relationship between initial accuracy and the ability to recognize errors is a key finding, indicating that reasoning confidence and self-awareness may not yet co-develop in current LLMs.

\subsection{Discussion}
The results of our study reveal three key insights:  
(1) \textbf{Culturally grounded reasoning remains challenging:} Even the top-performing models struggle to consistently produce correct answers, indicating that while LLMs can generate semantically relevant responses, they often miss the precise, creative insights required for riddle-solving.
(2) \textbf{Limited Gains from Few-Shot Prompting:} Adding random, contextual, or semantically similar exemplars provides only modest and inconsistent improvements, suggests that for tasks requiring a blend of creative reasoning and deep cultural knowledge. 
(3) \textbf{Performance and self-awareness are inversely related:} High-performing models, such as Gemini 2.5 Pro, show near-perfect confidence but often fail to recognize their own mistakes, while lower-performing models like LLaMA 4 Scout are more cautious and better at evaluating their answers. 

Overall, these results highlight persistent gaps in multilingual reasoning and emphasize the need for culturally grounded, reflective evaluation frameworks across languages.

\section{Conclusion}
This paper presented a systematic evaluation of large language models’ reasoning and self-assessment capabilities across seven major Indian languages using riddle-solving as a culturally grounded task. Our three-phase study-covering model selection, dataset curation, and dual-stage evaluation-showed that while models like Gemini 2.5 Pro achieved the highest accuracy, few-shot prompting offered only limited gains, and no single strategy proved consistently effective. Performance varied widely across languages, underscoring the ongoing challenges of multilingual reasoning. The self-evaluation results further reveal that stronger models tend to be overconfident, whereas weaker ones show greater awareness of their own mistakes. This inverse relationship suggests that current LLMs can generate likely answers but still struggle to check whether those answers are actually correct. Future work will expand this analysis to additional languages and reasoning types while exploring methods to enhance model self-awareness through reflective training.

\section{Ethics Statement}
The development of foundational LLMs has been overwhelmingly English-centric, a trend that risks creating a significant digital divide and marginalizing non-English native speakers. In a nation like India, where estimates suggest only 10-15\% of the population understands English, an over-reliance on Anglocentric models is an ethical concern.

For LLMs to be inclusive and beneficial tools, they must be able to grasp the cultural context, nuance, and implicit cues of diverse linguistic communities. This capability is essential for linguistic minorities and marginalized communities to feel represented and to interact safely with AI systems, rather than encountering "cultural and contextual misalignments".

This work aims to foster the development of foundational models that are not just multilingual, but truly multicultural in their reasoning capabilities, ensuring that the benefits of AI are accessible and relevant to all.

\section{Acknowledgements}

We sincerely thank Professor Dipti Misra Sharma for her insightful feedback, particularly the suggestion to conduct an Inter-Annotator Agreement study. We also extend our gratitude to R S Mughil Srinivasan and Kesavan T for their contributions to the Tamil riddle collection. Our thanks also go to the annotation teams across the seven Indian languages for their effort and dedication in curating and validating the multilingual riddle dataset.

\section{Bibliographical References}\label{sec:reference}

\bibliographystyle{lrec2026-natbib}
\bibliography{lrec2026-example}

% \label{lr:ref}
% \bibliographystylelanguageresource{lrec2026-natbib}
% \bibliographylanguageresource{languageresource}

\end{document}